\documentclass{article} 
\usepackage{iclr2027_conference,times}
\usepackage{enumitem}

\usepackage{amsmath,amsfonts,bm}

\def\eqref#1{equation~\ref{#1}}

\def\1{\bm{1}}

\DeclareMathAlphabet{\mathsfit}{\encodingdefault}{\sfdefault}{m}{sl}
\SetMathAlphabet{\mathsfit}{bold}{\encodingdefault}{\sfdefault}{bx}{n}

\usepackage{url}
\usepackage{amssymb}
\usepackage{booktabs}
\usepackage{colortbl}
\usepackage{tabularx}
\usepackage{wrapfig}
\usepackage{tikz}
\usetikzlibrary{arrows.meta,positioning}
\usepackage{tcolorbox}
\usepackage{float}
\usepackage{listings}
\usepackage{graphicx}
\usepackage{hyperref}
\usepackage{fontawesome5}

\definecolor{gaiaBlueLight}{RGB}{232,242,252}
\definecolor{gaiaGreenLight}{RGB}{232,247,237}
\definecolor{resourceBlue}{RGB}{30,64,175}

\title{Traverse: Learning When to Remember, Reset, and Redirect for Long-Horizon Web Search}

\newif\ifpreprint
\preprinttrue 

\author{
\small
Jingyuan Ma$^{1}$\thanks{Work done during an internship at ByteDance.},
Lynx Aster$^{2}$,
He Zhang$^{2}$,
Siyao Song$^{2}$,
Weijie Yuan$^{2}$,
Zhe Zhang$^{2}$,
Kai Jia$^{2}$\thanks{Corresponding authors.},
Zhifang Sui$^{1}$\footnotemark[2] \\
\\
\small $^{1}$State Key Laboratory of Multimedia Information Processing, 
\small School of Computer Science, Peking University \\
\small $^{2}$ByteDance \\
\small \texttt{mjy@stu.pku.edu.cn} \\[0.8ex]
\normalfont\footnotesize \faGithub\quad
\href{https://github.com/ByteDance-BandAI/Traverse}{\textcolor{resourceBlue}{\texttt{github.com/ByteDance-BandAI/Traverse}}} \\
\normalfont\footnotesize \raisebox{-0.22\height}{\includegraphics[height=1.25em]{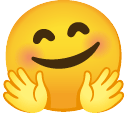}}\quad
\href{https://huggingface.co/datasets/ByteDance-BandAI/Traverse-AutoGen}{\textcolor{resourceBlue}{\texttt{huggingface.co/datasets/ByteDance/Traverse-AutoGen}}}
}

\hypersetup{
  pdftitle={Traverse: Learning When to Remember, Reset, and Redirect for Long-Horizon Web Search},
  pdfauthor={Jingyuan Ma, Lynx Aster, He Zhang, Siyao Song, Weijie Yuan, Zhe Zhang, Kai Jia, Zhifang Sui}
}

\ifpreprint
\iclrfinalcopy
\fi
\begin{document}

\maketitle
\ifpreprint
\lhead{Preprint}
\fi

\begin{abstract}
Long-horizon information-seeking agents often accumulate noisy or misleading context, causing early mistakes to persist and making recovery increasingly difficult. We introduce an autonomous search harness in which the agent manages its own search process through three states: Rubric, Answer, and Verify. The agent first defines criteria for a valid answer, searches under these criteria, and then independently verifies the result before deciding whether to terminate or continue searching. It is further equipped with a Seal Memory tool that enables active context management. Training this behavior with reinforcement learning, however, can induce Seal Collapse, resulting in unstable training and preventing the agent from reliably learning when and how to use its memory tools. We solve this with a simple strategy that trains only the final segment after context management. Our 35B model achieves \textbf{72.83 on BrowseComp}, outperforming comparable open-source systems, and consistently improves over the base model across BrowseComp-ZH, xbench, DeepSearchQA, WideSearch, financial investigation, and product search. Ablations show that autonomous compression outperforms automatic compaction and validate our RL design.
\end{abstract}

\section{Introduction}
Over the past two years, we have witnessed a remarkable transition of large language models from conversational assistants \citep{lambert2025tulu3} to increasingly autonomous agents that can interact with the environment \citep{wang2025openhands,zheng2025deepresearcher,li2025webthinker,wan2026deepverifier}. Powered by rapid advances in both model capabilities and the surrounding runtime infrastructure, or agent harnesses \citep{wu2024autogen,yang2024sweagent,wang2025openhands}, LLM agents are beginning to create tangible value in real-world workflows, most notably in coding \citep{yang2024sweagent,wang2025openhands,pan2025swegym} and deep research \citep{zheng2025deepresearcher,li2025webthinker,wan2026deepverifier}. Deep information-seeking systems such as OpenAI deep research \citep{openai2025deepresearch} and Gemini Deep Research \citep{citron2024gemini} already assist users across an unusually broad spectrum of tasks, from tracking down a half-remembered movie to conducting comprehensive literature reviews and producing market intelligence reports. In parallel, the community has developed increasingly rigorous benchmarks \citep{wei2025browsecomp,chen2026browsecompplus,du2026deepresearchbench,gao2026drarena,avraham2026dream} to probe the limits of these systems, evaluating not only whether an agent can retrieve relevant information, but also how deeply, broadly, and persistently it can explore the open web.

However, we observe a persistent failure mode in existing information-seeking agents. During long-horizon search, agents inevitably accumulate noisy, irrelevant, or misleading evidence in their context. Earlier mistakes can then bias subsequent decisions, causing the agent to revisit unproductive search paths and eventually enter a degenerate state from which recovery becomes increasingly difficult. This not only wastes valuable context budget, but also amplifies early errors throughout the trajectory. Recent work has begun to address this problem through explicit context management: some methods periodically summarize the growing interaction history~\citep{wu2026resumunlockinglonghorizonsearch}, while others reconstruct an evolving research state at each step~\citep{chen2026iterresearchrethinkinglonghorizonagents}. In broader long-horizon agent settings, recent work explores active memory curation and context folding~\citep{zhang2026memoryactionautonomouscontext,sun2025scalinglonghorizonllmagent}. Despite this progress, these methods focus primarily on maintaining a useful search state. We argue that an equally important capability remains largely overlooked: the agent itself may know whether the answer it has reached is actually correct.

In this paper, we seek the answer to the following question: \textit{Can we provide the right tools and workflow to enable an information-seeking agent to fully control its own search process and verify its answer?} To this end, we propose a highly autonomous search harness that transitions among three states: Rubric, Answer, and Verify. Given a question, the agent first enters the Rubric state to formulate a set of criteria that a correct answer must satisfy, and then carries these criteria into the Answer state to guide its search. During this process, we equip the agent with a self-managed context mechanism: the \textbf{Seal Memory tool}, which it may invoke at its own discretion to compress the accumulated context and carry forward only the information it considers useful. Once a candidate answer is found, the agent transitions to the Verify state, where it assumes the role of a verifier and re-examines the evidence supporting the answer. Crucially, the agent itself decides whether the answer is sufficiently verified or whether it should return to the Answer state and resume searching.

Training such an agent introduces an additional challenge: while we can distill the desired behaviors from strong teacher models, we also want the agent to learn through environmental feedback how to use its memory tools effectively with reinforcement learning. However, we find that the commonly adopted strategy of optimizing all context segments~\citep{wu2026resumunlockinglonghorizonsearch} leads to unstable training and a failure mode we term \textbf{Seal Collapse}, preventing the model from reliably learning when to invoke the Seal Memory Tool. We address this with a simple yet effective strategy that optimizes only the final segment after context management. Using this recipe, our \textbf{Traverse-35B} model achieves \textbf{72.83 on BrowseComp}~\citep{wei2025browsecomp}, outperforming recent open-source systems of comparable scale, including QUEST-35B and AREX-Turbo~\citep{xie2026questtrainingfrontierdeep,lu2026arexrecursivelyselfimprovingagent}, while also showing consistent gains over the base model on representative benchmarks such as BrowseComp-ZH~\citep{zhou2025browsecompzhbenchmarkingwebbrowsing} and WideSearch~\citep{wong2025widesearchbenchmarkingagenticbroad}, with strong transfer to financial investigation and product search. Extensive ablations further show that agent-controlled context compression outperforms automatic compaction and validate the effectiveness of our RL training strategy. Our contributions can be summarized as follows:

\begin{itemize}[leftmargin=*]
    \item
    We develop an autonomous long-horizon search harness that combines a Rubric--Answer--Verify state machine with agent-triggered Seal and Read Memory tools. The agent decides when to create a context boundary, what evidence and failed hypotheses to preserve, and whether a candidate answer should terminate the search or trigger another evidence-seeking round.

    \item
    We identify \emph{Seal Collapse}, a failure mode that arises when trajectory-level feedback is propagated across all context segments, and analyze its connection to ambiguous credit assignment and segment-induced trajectory reweighting. We introduce a final-segment-only RL strategy that stabilizes memory-tool learning while keeping the training workload independent of the number of context resets.

    \item
    We demonstrate strong performance across a broad range of search tasks, together with consistent improvements over the base model. Controlled ablations further validate the effectiveness of agent-controlled context compression and final-segment-only RL. 
\end{itemize}

\section{Method}
\subsection{Overview}
\label{sec:overview}
\begin{figure*}[t]
    \centering
    \includegraphics[width=\textwidth]{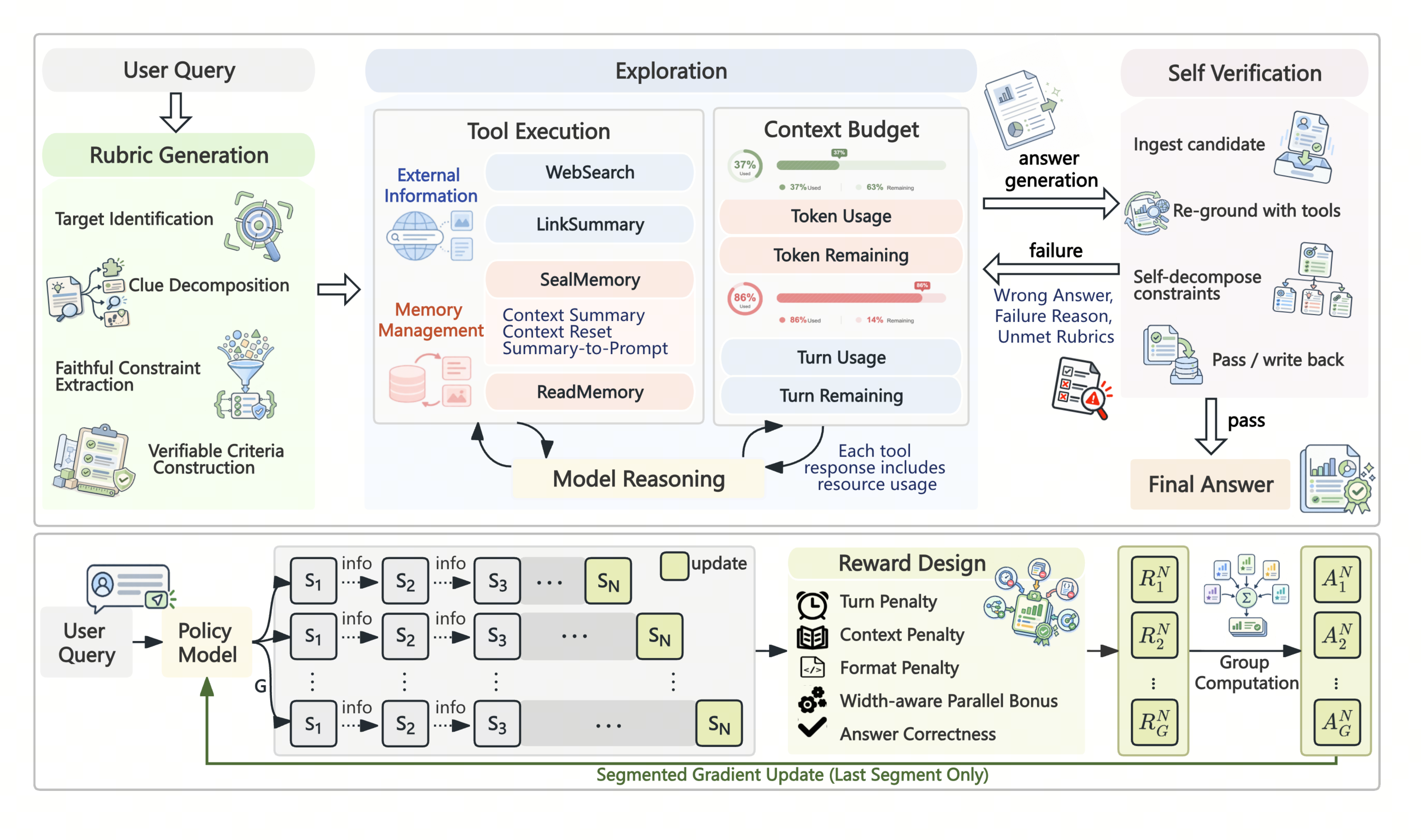}
    \caption{\textbf{Overview of Traverse.}
    \textit{Top:} Traverse integrates rubric-guided exploration, autonomous context management, and self-verification into an iterative search workflow.
    \textit{Bottom:} During RL, trajectories are segmented by Seal Memory operations, and only the final segment is optimized using group-relative advantages.}
    \label{fig:framework_overview}
\end{figure*}

Our goal is to maximize agent autonomy in long-horizon search. We first address the unavoidable problem of context exhaustion by letting the agent decide both when to compress its context and what information to carry forward. This capability is implemented through two tools: the \textbf{Seal Memory tool}, which compresses and stores memory, and the \textbf{Read Memory tool}, which retrieves finer-grained details when needed. We further organize the search process as a state machine. Given a query $q$, the agent first enters the \textsc{Rubric state} and constructs $R=\{r_i\}_{i=1}^m$. These rubrics are then passed to the \textsc{Answer state}, where the agent gathers evidence with search and browse tools while autonomously deciding whether to compress its context or produce an answer $y$. In the \textsc{Verify state}, the agent evaluates $y$ conditioned on $q$ and $R$, and outputs $d\in \{\text{Pass},\text{Revise Answer}\}$. The process terminates on Pass; otherwise, the agent returns to the Answer state with verifier feedback and continues searching. Our framework is summarized in Figure~\ref{fig:framework_overview}.

\subsection{Self-Context Management}
\label{sec:context}
Context management is unavoidable in long-horizon agentic search. When an agent follows an incorrect search path, or must verify an answer across multiple pieces of evidence, its finite context can easily become exhausted. We observe, however, that the agent itself often has sufficient signal to recognize when search has stalled and when the evidence is strong enough to discard accumulated noise. We therefore introduce the Seal Memory Tool, an agent-invoked context management mechanism that lets the agent autonomously decide whether to continue searching or reset its context.
When invoking the tool, the agent is required to summarize the information worth preserving according to a structured memory template. It reviews its interaction history, incorporates evidence collected so far, and, if available, the memory produced by the previous segment. We expose several summary fields as tool arguments to guide this process. Let $h_k$ denote the interaction history in segment $k$, and $m_k$ the memory carried into that segment. The next memory is generated as
\begin{equation}
    m_{k+1} \sim \pi_\theta(\cdot \mid q, h_k, m_k)
\end{equation}
with $m_0=\varnothing$. After sampling $m_{k+1}$, the context is reset and the next segment starts from
\begin{equation}
    c_{k+1}=q \oplus m_{k+1}
\end{equation}
The agent then resumes search from $c_{k+1}$. The trajectory terminates once the agent chooses to produce a final answer instead of invoking the memory tool again. Meanwhile, after each tool call, the agent is informed of its current token usage, allowing it to explicitly track the remaining context budget rather than blindly continuing until it hits the context limit. We also provide a Read Memory Tool to retrieve detailed information from the stored memory $m_k$ when necessary. More details about the memory tools are provided in Appendix~\ref{app:tool-schemas}.

\subsection{Deep Research State Machine}
\label{sec:verify}
In this section, we explore how to equip the agent with self-verification capabilities. Inspired by DeepSeekMath-V2~\citep{shao2025deepseekmathv2}, we design a workflow that spans question decomposition through answer verification, enabling the agent to assess the correctness of its own answer and decide whether the search process should terminate.

\noindent\textbf{Rubric State.} We first ask the agent to decompose the question in the \textsc{Rubric State}. Given a query \(q\), the agent identifies the criteria that must be satisfied for an answer to be considered correct and produces a rubric set \(R=\{R_i\}_{i=1}^{n}\), where the number of criteria \(n\) is determined adaptively by the agent. The rubric turns the constraints in the query into an explicit, structured checklist that can guide subsequent search and verification. For the multi-constraint, short-answer tasks considered in this work, \(R\) typically contains one content criterion for each distinct clue in the question, together with an additional criterion specifying the required answer format. Each content criterion describes an independently verifiable property of the target entity and is designed to be checked against Web evidence. To avoid injecting assumptions at this early stage, the criteria remain simple, objective, and faithful to the original query, preserving approximate expressions, numerical ranges, and indirect descriptions at their original level of specificity. We represent each criterion as an indexed natural-language description and carry the resulting rubric forward to subsequent states.

\noindent\textbf{Answer State.} 
Next, the agent enters the \textit{Answer State} and begins the actual search. Given the rubric set $R$ produced in the previous state, the agent is equipped with search, Web browsing, Seal Memory, and Read Memory tools. Each tool response is augmented with the current resource usage:
\begin{tcolorbox}[
    width=\linewidth,
    colback=gray!10,
    colframe=gray!70!black,
    boxrule=0.6pt,
    arc=2mm,
    left=6pt,
    right=6pt,
    top=5pt,
    bottom=5pt
]
\textless token\_budget\textgreater{} Used: x / Total: y; Remain: y-x \textless/token\_budget\textgreater\\
\textless turn\_budget\textgreater{} Used: a / Total: b; Remain: b-a \textless/turn\_budget\textgreater
\end{tcolorbox}
Based on the remaining budget and the noisiness of the accumulated context, the agent decides whether to invoke the Seal Memory Tool to compact its context. Within each segment, it repeatedly chooses between continuing the search through compaction or terminating with an answer $y$. When compaction resets the context, the token budget is refreshed, while the turn budget is preserved across segments.

\noindent\textbf{Verification State.}
Finally, the agent enters the \textit{Verify State}, where it receives the query $q$, the candidate answer $y$, and the rubric set $R$, and independently assesses whether $y$ satisfies the required criteria. We equip the verifier with the same four tools as the Answer State---search, Web browsing, Seal Memory, and Read Memory---allowing it to gather external evidence for each rubric. When the verification context becomes overly long or noisy, the verifier may also invoke Seal Memory to compact its context. At the end of verification, the agent outputs a decision $d\in\{\textsc{Pass},\textsc{ReviseAnswer}\}$. If $d=\textsc{Pass}$, the process terminates and $y$ is returned as the final answer. Otherwise, the verifier produces a suggestion explaining why the current answer is insufficient, which rubrics remain unsatisfied, and where the subsequent search should focus, before transitioning back to the Answer State. Upon re-entry, the Answer State is additionally informed of previously proposed answers and their failures, together with the verifier's latest feedback and unmet rubrics, enabling the agent to resume search with a more targeted direction rather than starting from scratch.

Taken together, the agent cycles through the Rubric, Answer, and Verify states, potentially undergoing multiple rounds of answering and verification until it can no longer identify an error and decides to submit the answer. Throughout this process, the agent autonomously manages its context via the Seal Memory Tool and independently determines whether the current answer is sufficiently verified or whether further search is necessary.

\section{Model Training}

\subsection{Supervised Fine-Tuning}

\noindent\textbf{Query Construction and Data Curation.} To train our model, we first synthesize challenging yet verifiable information-seeking questions using knowledge graphs, following WebShaper~\citep{tao2025webshaperagenticallydata}, and apply rigorous verification and cleaning to retain only solvable instances. We then pair these questions with a strong teacher model and our harness to collect agent trajectories. Since teacher rollouts can still contain undesirable behaviors, we filter both rule violations, such as malformed or unparsable tool calls, and behavioral errors, such as invoking the Seal Memory Tool prematurely or after the answer has already been found. Rather than discarding an entire trajectory due to a few flawed turns, we mask those turns out during training while preserving the remaining valid supervision. Further details about data synthesis are provided in Appendix~\ref{app:task-synthesis}.

\noindent\textbf{Masked SFT.} To incorporate the filtering masks above, we augment the standard next-token prediction objective with a turn-level training mask. Specifically, for each turn $i$, we assign a mask $\mathcal{M}_i\in\{0,1\}$, which is broadcast to all tokens in that turn. All system, user, and tool turns are always assigned $\mathcal{M}_i=0$. For assistant turns, $\mathcal{M}_i=0$ if the turn is identified as undesirable and $1$ otherwise. The SFT objective becomes
\begin{equation}
\mathcal{L}_{\mathrm{SFT}} =
-\frac{1}{\sum_{i=1}^{N}\mathcal{M}_i T_i}
\sum_{i=1}^{N}
\mathcal{M}_i
\sum_{j=1}^{T_i}
\log \pi_\theta\left(x_{i,j}\mid x_{<i},x_{i,<j}\right)
\end{equation}
where $N$ is the number of turns, $T_i$ is the number of tokens in turn $i$, and $x_{i,j}$ denotes its $j$-th token. This allows the agent to avoid imitating bad patterns while still learning how to recover after it.

\subsection{Seal Guided Reinforcement Learning}
\label{sec:seal-rl}




After distilling context-management behaviors from strong-teacher trajectories, we further ask whether the agent can learn them from environmental feedback. Context compaction introduces a structural challenge for RL: every context reset fragments the original trajectory into a new training sample. A trajectory $\tau_i$ may therefore contain multiple segments $\tau_i=\{\tau_i^{(1)},\ldots,\tau_i^{(K_i)}\}$, where $K_i$ is the number of segments induced by compaction.

A natural strategy is to assign each trajectory its final correctness as reward $r_i$, compute the group-normalized advantage $\hat A_i=(r_i-\mu_{\mathcal G})/\sigma_{\mathcal G}$, where $\mu_{\mathcal G}$ and $\sigma_{\mathcal G}$ are the group reward mean and standard deviation, and broadcast it to every segment and optimized token: $\hat A_{i,k,t}=\hat A_i$. This follows the philosophy of GRPO, where all optimized tokens in a trajectory share the same advantage.

This seemingly natural choice creates two problems. \textbf{First}, it obscures credit assignment. A successful trajectory may recover only after its final compaction removes misleading evidence from earlier exploration. Conversely, a failed trajectory may contain useful evidence in earlier segments yet make a wrong commitment only in the last. Broadcasting the final outcome therefore rewards or penalizes many actions only weakly related to the result, injecting substantial optimization noise. \textbf{Second}, fragmentation implicitly reweights trajectories. The RL objective can be written as
\begin{equation}
\mathcal L_{\mathrm{RL}}=\frac{1}{\sum_{i=1}^{N}K_i}\sum_{i=1}^{N}\sum_{k=1}^{K_i}\mathcal L\left(\pi_\theta, \tau_i^{(k)},\hat A_i\right)
\end{equation}
where $N$ is the number of trajectories, $\mathcal{L}$ is an RL objective such as PPO, and $\pi_\theta$ is the policy parameterized by $\theta$. Suppose one trajectory is split into $K$ segments while the remaining $N-1$ trajectories are not compacted. The batch now contains $K+N-1$ training samples. Each non-compacted trajectory receives weight $1/(K+N-1)$ instead of $1/N$; thus, the normalizer changes with $K$.

Addressing both issues simultaneously is non-trivial. One must either spend additional computation to obtain more accurate value estimates, thereby enabling finer-grained credit assignment, or resort to more sophisticated algorithmic designs whose effectiveness must then be carefully validated under this dynamically evolving training process. Instead, we take a deliberately simple route. We propose a simple yet effective strategy that sidesteps both challenges: we optimize only the final segment produced by context compaction. Despite its simplicity, this design is motivated by two key observations:



\begin{enumerate}[leftmargin=*]
    \item \textbf{Implicit Learning.} At the end of each segment, the agent must either produce an answer or invoke the Seal Memory Tool, otherwise, the segment terminates due to context overflow and becomes the final segment, where failure is penalized. Thus, even without explicitly rewarding memory tool calls, the agent is implicitly encouraged to invoke the Seal Memory Tool when further search is needed, while learning when to stop and answer within the current segment.
    \item \textbf{Reward Signal.} The final segment is temporally closest to the trajectory-level reward and therefore receives a less noisy learning signal, reducing the optimization variance induced by broadcasting the same advantage across all preceding segments.
\end{enumerate}
In this way, training only the final segment actually couples two decisions: whether to continue searching through compaction, and whether enough evidence has been gathered to answer. We therefore formulate our RL objective as:
\begin{equation}
\mathcal L_{\mathrm{RL}}^{\mathrm{last}}=\frac{1}{N}\sum_{i=1}^{N}\mathcal L\left(\pi_\theta, \tau_i^{(K_i)},\hat A_i\right)
\end{equation}
where only the last segment $\tau_i^{(K_i)}$ is optimized. We compute within-group advantages using the GRPO-style Z-score normalization. For policy optimization, we adopt GSPO~\citep{zheng2025gspo} to stabilize MoE training. Specifically, GSPO aggregates token-level policy ratios into a sequence-level importance ratio, $\rho_i=\exp\left(\frac{1}{T_i}\sum_{j=1}^{T_i}\log\frac{\pi_\theta(x_{i,j}\mid x_{<i},x_{i,<j})}{\pi_{\theta_{\mathrm{old}}}(x_{i,j}\mid x_{<i},x_{i,<j})}\right)$. We additionally apply penalties for tool usage, trajectory length, and token-budget consumption. Detailed reward configurations are provided in Appendix~\ref{app:reward-shaping}.

\section{Experiments}
\subsection{Experiment Setup}
\noindent\textbf{Implementation Details.}
We use Qwen3.5-35B-A3B as our base model and adopt GSPO to stabilize RL training for the MoE architecture. We set the maximum number of agent turns to 512, the context length to 256K, and the sampling temperature to 1.0. Our RL training uses 96 GPUs, with 32 GPUs allocated to policy optimization and 64 GPUs dedicated to asynchronous rollout generation. For all ablation studies, we keep the search backend, training hyperparameters, and sampling configuration identical to ensure fair comparisons. During SFT, we collect and train on trajectories from the rubric, answer, and verification states, whereas RL is applied only to the answer state. For BrowseComp, we additionally enable a Discard-All context reset when the context is exhausted, following common baseline practice; this fallback is orthogonal to Seal Memory and can be combined with the agent-triggered compression mechanism.

\noindent\textbf{Baselines.} We compare our method with the closed-source models like GPT-5.5~\citep{openai2026gpt55}, Claude Opus 5~\citep{anthropic2026opus5}, and Gemini 3.1 Pro~\citep{googledeepmind2026gemini31pro}, as well as the open-source models like GLM-5.1~\citep{glm5team2026glm5}, Kimi K3~\citep{kimiteam2026kimik3openfrontier}, DeepSeek-V4-Pro~\citep{deepseekai2026deepseekv4highlyefficientmilliontoken}, Qwen3.5-35B-A3B and Qwen3.5-122B-A10B~\citep{qwen35blog}, MiroThinker-1.7-mini~\citep{miromindteam2026mirothinkerpushingperformanceboundaries}, QUEST-35B~\citep{xie2026questtrainingfrontierdeep}, FORT-Searcher~\citep{deng2026fortsearchersynthesizingshortcutresistantsearch}, and AREX-Turbo~\citep{lu2026arexrecursivelyselfimprovingagent}.

\noindent\textbf{Benchmarks.} We evaluate our model on five search benchmarks: \textbf{BrowseComp} for deep information seeking, \textbf{BrowseComp-ZH (BC-ZH)}~\citep{zhou2025browsecompzhbenchmarkingwebbrowsing} for multilingual search, \textbf{WideSearch}~\citep{wong2025widesearchbenchmarkingagenticbroad} for breadth-oriented retrieval, and \textbf{xbench-DeepSearch (xbench)}~\citep{chen2025xbenchtrackingagentsproductivity} and \textbf{DeepSearchQA (DSQA)}~\citep{gupta2026deepsearchqabridgingcomprehensivenessgap} for multi-step evidence collection and synthesis. Together, they cover diverse deep-search capabilities across languages and task formats. We further evaluate generalization on \textbf{GAIA}~\citep{gaia}, \textbf{FinSearchComp}~\citep{hu2025finsearchcomprealisticexpertlevelevaluation}, and \textbf{ShoppingComp}~\citep{tou2025shoppingcomp}. Additional benchmark and evaluation details are provided in Appendix~\ref{app:judge_settings}.

\subsection{Main Results}
\begin{table*}[t]
\centering
\caption{Main results on five search-agent benchmarks. We report accuracy on BrowseComp, BrowseComp-ZH, and xbench-DeepSearch-2510, Macro-F1 on DeepSearchQA, and Item-F1 on WideSearch. ``--'' denotes unavailable results.}
\label{tab:main_results}
\small
\begingroup
\setlength{\tabcolsep}{1.5pt}
\renewcommand{\tabularxcolumn}[1]{m{#1}}
\begin{tabularx}{\linewidth}{>{\hsize=1.7\hsize\raggedright\arraybackslash}X*{5}{>{\hsize=.86\hsize\centering\arraybackslash}X}}
\toprule
\textbf{Model}
& \shortstack[c]{\textbf{Browse}\\\textbf{Comp}}
& \shortstack[c]{\textbf{BrowseComp-}\\\textbf{ZH}}
& \shortstack[c]{\textbf{xbench-}\\\textbf{2510}}
& \shortstack[c]{\textbf{DeepSearch}\\\textbf{QA}}
& \shortstack[c]{\textbf{Wide}\\\textbf{Search}} \\
\midrule

\rowcolor{gaiaBlueLight}
\multicolumn{6}{l}{\textit{Closed-source Models}} \\

\rowcolor{gaiaBlueLight}
GPT-5.5
    & 84.4 & -- & -- & -- & -- \\

\rowcolor{gaiaBlueLight}
Gemini 3.1 Pro
    & 85.9 & -- & 53.0 & 93.3 & 66.4 \\

\rowcolor{gaiaBlueLight}
Claude Opus 5
    & 90.8 & -- & -- & 95.0 & -- \\

\midrule

\rowcolor{gaiaGreenLight}
\multicolumn{6}{l}{\textit{General-purpose Models}} \\

\rowcolor{gaiaGreenLight}
Qwen3.5-35B-A3B
    & 61.0 & 69.5 & 50.3 & 68.5 & 57.1 \\

\rowcolor{gaiaGreenLight}
Qwen3.5-122B-A10B
    & 63.8 & 69.9 & -- & -- & 60.5 \\

\rowcolor{gaiaGreenLight}
GLM-5.1
    & 79.3 & -- & -- & -- & -- \\

\rowcolor{gaiaGreenLight}
DeepSeek-V4-Pro
    & 83.4 & -- & 80.0 & 88.7 & 78.0 \\

\rowcolor{gaiaGreenLight}
Kimi K3
    & 91.2 & -- & -- & 95.0 & -- \\

\midrule

\rowcolor{gaiaGreenLight}
\multicolumn{6}{l}{\textit{Search-specialized Models}} \\

\rowcolor{gaiaGreenLight}
IterResearch-30B-A3B
    & 37.3 & 45.2 & -- & -- & -- \\

\rowcolor{gaiaGreenLight}
REDSearcher
    & 42.1 & 49.8 & -- & -- & -- \\

\rowcolor{gaiaGreenLight}
QUEST-35B
    & 64.6 & -- & -- & -- & 60.6 \\

\rowcolor{gaiaGreenLight}
MiroThinker-1.7-mini
    & 67.9 & 72.3 & 57.2 & 67.9 & -- \\

\rowcolor{gaiaGreenLight}
AREX-Turbo
    & 70.7 & -- & 57.0 & 78.5 & 68.5 \\

\rowcolor{gaiaGreenLight}
FORT-Searcher
    & 72.2 & 75.0 & 57.2 & -- & -- \\

Traverse-35B (Ours)
    & \textbf{72.8}
    & \textbf{70.7}
    & \textbf{57.0}
    & \textbf{82.8}
    & \textbf{72.1} \\

\bottomrule
\end{tabularx}
\endgroup
\end{table*}


In this section, we evaluate our model across a diverse suite of benchmarks and compare it against strong existing systems. Beyond widely used search benchmarks, we further probe its out-of-domain generalization on finance and e-commerce-oriented tasks. As shown in Table~\ref{tab:main_results}, Traverse-35B achieves a score of \textbf{72.83 on BrowseComp}, reaching performance comparable to FORT-Searcher. Although it trails slightly on BrowseComp-ZH and xbench, it remains highly competitive. More notably, on DeepSearchQA and WideSearch, our model achieves state-of-the-art performance among search-specialized models. Table~\ref{tab:domain_results} further highlights the robustness of this capability beyond the training distribution: on both financial and e-commerce benchmarks, Traverse-35B delivers substantial gains over the base model, suggesting that the learned search behaviors transfer effectively to previously unseen domains. Our method improves GAIA accuracy by 18.45 percentage points, FinSearchComp accuracy by 19.70 percentage points, and ShoppingComp SoP by 0.1331 over the base model.

\begin{table}[H]
\centering
\caption{Generalization results on GAIA, FinSearchComp, and ShoppingComp. VPR measures the valid-response rate, while SoP denotes Satisfaction of Products.}
\label{tab:domain_results}
\small
\setlength{\tabcolsep}{10pt}
\begin{tabular}{@{}lcccc@{}}
\toprule
& \textbf{GAIA}
& \textbf{FinSearchComp}
& \multicolumn{2}{c}{\textbf{ShoppingComp}} \\
\cmidrule(lr){2-2}
\cmidrule(lr){3-3}
\cmidrule(lr){4-5}
\textbf{Model}
& \textbf{Acc. (\%)}
& \textbf{Acc. (\%)}
& \textbf{VPR}
& \textbf{SoP} \\
\midrule
Base
    & 61.81 & 38.36 & 0.6660 & 0.2245 \\
\textbf{Ours}
    & 80.26 & 58.06 & 0.6680 & 0.3576 \\
\bottomrule
\end{tabular}
\end{table}

\subsection{Ablation Studies}
In this section, we investigate the effectiveness of our approach from three perspectives. We first validate the proposed RL algorithm, then quantify the gains brought by RL over the SFT model, and finally compare our method against Auto Compaction. For BrowseComp, we evaluate on a 185-question subset to reduce evaluation cost, which we refer to as \textbf{BC185}. Further details about BC185 are provided in Appendix~\ref{app:browsecomp-lite}.

\begin{figure*}[t]
    \centering
    \includegraphics[width=\textwidth]{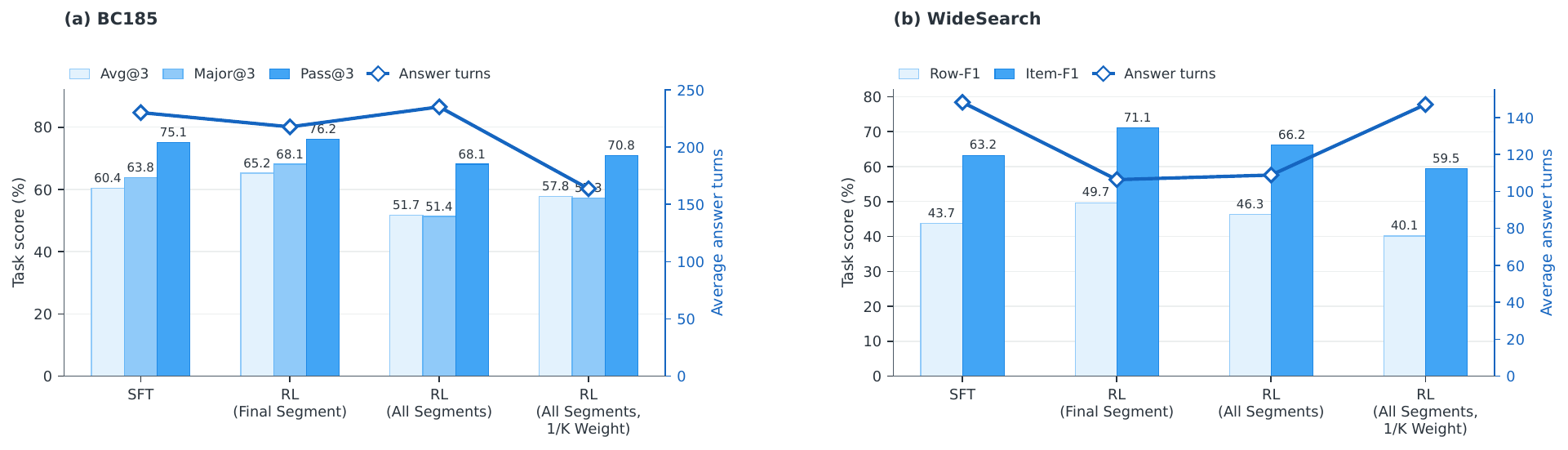}
    \caption{Comparison of SFT, final-segment RL, all-segment RL, and all-segment RL with $1/K_i$ weighting on BC185 and WideSearch. In the weighted variant, each of the $K_i$ segments in trajectory $i$ is assigned weight $1/K_i$, so every trajectory has the same total weight. Bars report task performance on the left axis, and lines report average answer turns on the right axis, where lower is better. All evaluations are conducted with answer verification disabled.}
    \label{fig:bc185_widesearch_comparison}
\end{figure*}

\noindent\textbf{Is Training the Last Segment Effective?}
We compare final-segment RL (Section~\ref{sec:seal-rl}) with two alternatives that broadcast the trajectory-level advantage to all segments: standard all-segment RL and a $1/K_i$-weighted variant that equalizes total trajectory weights. As shown in Figure~\ref{fig:bc185_widesearch_comparison}, standard all-segment RL reduces BC185 Avg@3 from \textbf{60.4} for SFT to \textbf{51.7}. Weighting partially recovers it to \textbf{57.8}, still below SFT and final-segment RL (\textbf{65.2}). On WideSearch, the weighted variant achieves \textbf{40.1}/\textbf{59.5} Row-F1/Item-F1, compared with \textbf{46.3}/\textbf{66.2} for standard all-segment RL and \textbf{49.7}/\textbf{71.1} for final-segment RL. Thus, correcting trajectory reweighting alone does not resolve the degradation from all-segment training.

\begin{figure*}[t]
    \centering
    \includegraphics[width=\textwidth]{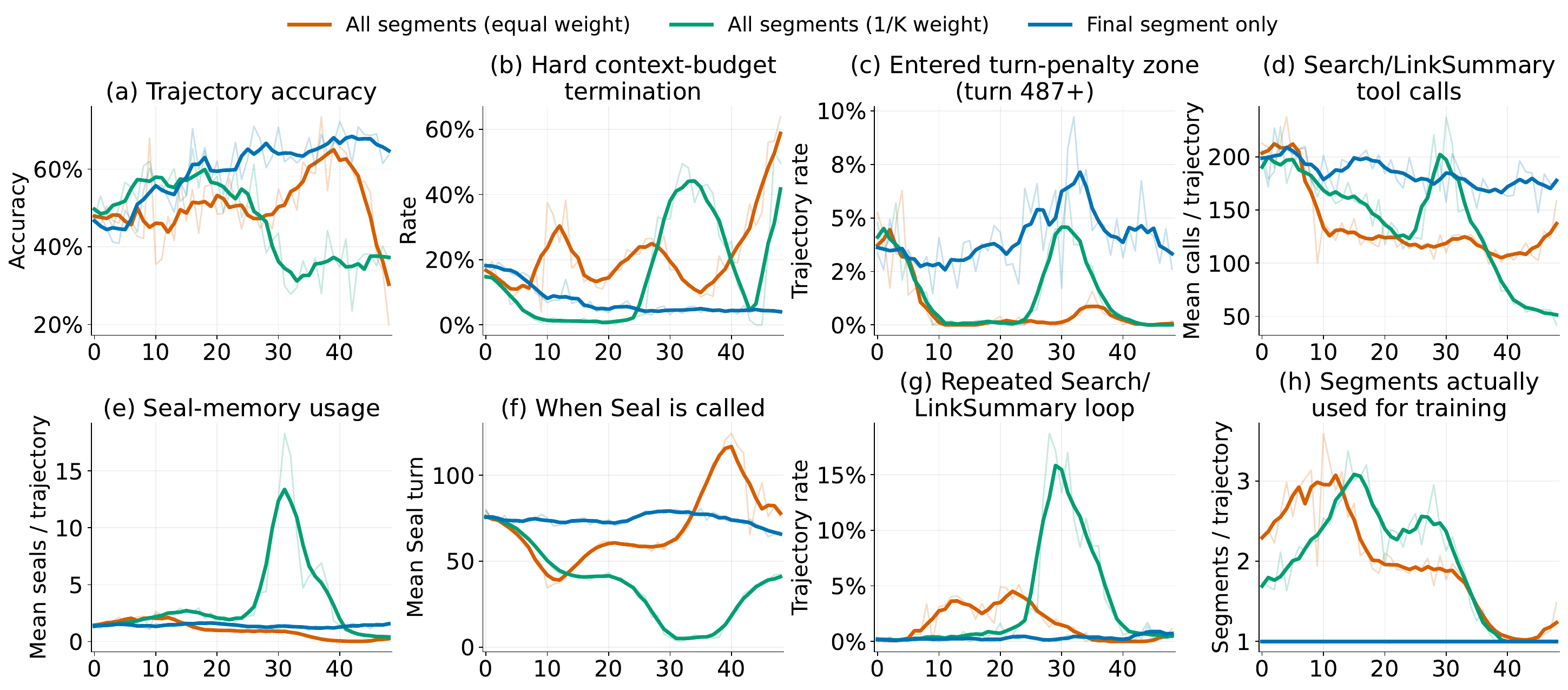}
    \caption{Training dynamics of equal-weight all-segment RL, $1/K_i$-weighted all-segment RL, and final-segment-only RL. The panels report trajectory accuracy, hard context-budget termination, entry into the turn-penalty zone, Search/LinkSummary tool calls, Seal Memory usage and timing, repeated Search/LinkSummary loops, and the number of segments used for training.}
    \label{fig:main_ablation_dynamics}
    \vspace{-12pt}
\end{figure*}

Figure~\ref{fig:main_ablation_dynamics} reveals distinct training failures. Standard all-segment RL initially improves but later collapses: Seal calls are delayed, usage approaches zero, and hard context-budget termination rises sharply. The $1/K_i$-weighted variant shows accuracy degradation from around step 20, followed by a surge in Seal usage around steps 25--30, increasingly early Seal calls, and repeated Search/LinkSummary loops. These patterns are consistent with unresolved credit-assignment difficulties despite trajectory reweighting. Final-segment RL maintains stable accuracy, Seal usage, and search behavior while optimizing only one segment per trajectory, avoiding growth in training-segment count as context resets increase.

\noindent\textbf{How Much Does RL Improve over SFT?}
Here, we quantify the gains brought by RL over SFT. We compare the SFT and RL models on BC185 and WideSearch. On BC185, RL improves the score by \textbf{4.8 points} while reducing the number of agent turns required to reach the answer. We observe the same trend on WideSearch, where RL improves Row F1 by \textbf{6.0 points} and Item F1 by \textbf{7.9 points}, again with fewer search turns. These results show that RL yields substantial improvements in both effectiveness and search efficiency, while further validating the effectiveness of optimizing only the final segment.

\noindent\textbf{Can Seal Memory Outperform Auto Compaction?}
We further compare our proposed Seal Memory mechanism with the widely used Auto Compaction strategy. We implement an Auto Compaction harness that triggers summarization when token usage reaches 90\% of the context window, i.e., roughly 230K tokens in a 256K context. For a fair comparison, the summarizer is the same model used by the main agent: once the threshold is reached, it receives the full interaction history and produces a compact memory in the same format as the Seal Memory Tool, after which the agent continues from the user prompt augmented with this memory. As shown in Table~\ref{tab:active_vs_compaction}, allowing the agent to decide when to compact its context yields a substantial performance gain: Active Seal improves Avg@3 by \textbf{10.28 points}, with consistent gains on Pass@3 and Maj@3. More importantly, it reorganizes the search process much earlier, with the median first Seal occurring at 67 turns and 67.5K tokens, compared with 217 turns and 231.4K tokens for Auto Compaction. The flexibility of Seal Memory enables the agent to explore a broader range of possibilities and pursue more diverse search paths. Its average trajectory length increases from 112.2 to 217 turns, ultimately achieving substantially higher final-answer accuracy than fixed-threshold compaction.

\begin{table*}[t]
\centering
\small
\caption{Active sealing versus automatic context compaction on BC185. \textbf{First Turn} and \textbf{First Tokens} denote the median turn index and token usage at the first memory operation, respectively, computed over trajectories containing a memory operation. Finally, \textbf{Avg. Turns} denotes the average number of turns across all trajectories.}
\label{tab:active_vs_compaction}
\begingroup
\setlength{\tabcolsep}{3pt}
\renewcommand{\tabularxcolumn}[1]{m{#1}}
\begin{tabularx}{\linewidth}{l*{6}{>{\centering\arraybackslash}X}}
\toprule
Strategy
& Avg@3
& Pass@3
& Maj@3
& \shortstack{First\\Turn}
& \shortstack{First\\Tokens}
& \shortstack{Avg.\\Turns} \\
\midrule
Auto Compaction
    & 54.95
    & 68.11
    & 56.22
    & 217
    & 231.4K
    & 112.2 \\
Active Seal
    & \textbf{65.23}
    & \textbf{76.22}
    & \textbf{68.11}
    & 67
    & 67.5K
    & 217.7 \\
\bottomrule
\end{tabularx}
\endgroup
\end{table*}

\section{Related Work}

\label{sec:related-work}

\subsection{Web Search Agents}

Large language models have increasingly been equipped with search and browsing tools to acquire external information during reasoning. Early systems such as WebGPT~\citep{nakano2022webgptbrowserassistedquestionansweringhuman} and ReAct~\citep{yao2023react} established the foundations for grounded answer generation and interleaved reasoning and action. Recent search agents extend this paradigm to longer and more autonomous interactions involving query reformulation, webpage navigation, evidence aggregation, and answer synthesis~\citep{li2025webthinker, wu2025webdancerautonomousinformationseeking, li2025websailornavigatingsuperhumanreasoning}. Meanwhile, benchmarks such as GAIA~\citep{gaia}, BrowseComp~\citep{wei2025browsecomp}, BrowseComp-ZH~\citep{zhou2025browsecompzhbenchmarkingwebbrowsing}, WideSearch~\citep{wong2025widesearchbenchmarkingagenticbroad}, and DeepSearchQA~\citep{gupta2026deepsearchqabridgingcomprehensivenessgap} evaluate complementary aspects of tool use, persistent browsing, multilingual retrieval, and broad information collection. Our framework organizes search into explicit \textsc{Rubric}, \textsc{Answer}, and \textsc{Verify} states, enabling the agent to construct task-specific evaluation criteria, conduct evidence-seeking interactions, and revise its answer according to verification outcomes.

\subsection{Reinforcement Learning for Long-Horizon Search Agents}
Recent work applies outcome-based reinforcement learning to teach language models when and how to search, allowing search behavior to emerge without intermediate reasoning annotations~\citep{jin2025searchr1trainingllmsreason,song2025r1searcherincentivizingsearchcapability,chen2025researchlearningreasonsearch}. Extending this approach to long-horizon search introduces two coupled challenges: the interaction history may exceed the active context budget, and a terminal reward provides ambiguous supervision for the many exploratory actions preceding the final answer. Context summarization and folding methods address the first challenge by compressing interaction histories and restructuring trajectories around context updates~\citep{wu2026resumunlockinglonghorizonsearch,zhang2026memoryactionautonomouscontext,sun2025scalinglonghorizonllmagent}. However, trajectory-level advantages are commonly propagated across multiple reconstructed segments, even though early search often contains irrelevant retrievals, candidate elimination, and abandoned directions. Our method uses autonomous \textsc{Seal Memory} operations to define segment boundaries and applies the terminal learning signal only to the final segment. Earlier evidence remains available through the sealed memory, while uncertain exploratory segments receive no direct gradient.

\section{Conclusion}
This paper introduces \textsc{Traverse}, a long-horizon search agent that autonomously manages both its search process and its context. The agent structures research through \textsc{Rubric}, \textsc{Answer}, and \textsc{Verify} states, while the Seal Memory Tool allows it to decide when to compress accumulated context and what information to preserve. To train this behavior, we combine masked supervised fine-tuning on curated teacher trajectories with a reinforcement-learning strategy that optimizes only the final segment after context compaction. This simple design avoids the noisy credit assignment and trajectory reweighting induced by all-segment training, preventing the Seal Collapse observed in our experiments. Using this recipe, \textsc{Traverse}-35B achieves 72.83 on BrowseComp and remains competitive across multilingual, broad-retrieval, and evidence-synthesis benchmarks, while transferring effectively to financial and e-commerce search tasks. Our ablations further show that final-segment training improves the performance, and that agent-triggered sealing substantially outperforms fixed-threshold automatic compaction. 

\bibliography{iclr2027_conference}
\bibliographystyle{iclr2027_conference}

\newpage
\appendix
\section{Search Task Synthesis}
\label{app:task-synthesis}

Our task-synthesis pipeline follows a formalization-driven design inspired by WebShaper~\citep{tao2025webshaperagenticallydata}.
It consists of four stages: seed-task initialization, structured formalization, Web-grounded expansion, and quality filtering.

\paragraph{Seed Selection and Formalization.}
We begin by sampling a target entity $x^\star$ from a large-scale knowledge graph.
A controlled traversal of its relational neighborhood retrieves a collection of factual relations and attributes, from which we construct a simple seed question with a known answer.
The target may be either the sampled entity itself or one of its attributes, while the remaining facts serve as initial constraints.

For each constraint, we define a knowledge projection
\begin{equation}
    \mathcal{C}_j
    =
    R_j(c_j)
    =
    \left\{
        x \mid R_j(x,c_j)
    \right\},
\end{equation}
where $R_j$ denotes a relation and $c_j$ is an entity or attribute value.
The projection $\mathcal{C}_j$ contains all candidate entities satisfying the corresponding constraint.
We denote the formal representation of the seed task as
\begin{equation}
    \Phi^{(0)}(x)
    =
    \bigwedge_{j=1}^{m} R_j(x,c_j),
\end{equation}
and define its feasible answer set as
\begin{equation}
    \operatorname{Ans}\!\left(\Phi^{(0)}\right)
    =
    \left\{
        x \mid \Phi^{(0)}(x)
    \right\}
    =
    \bigcap_{j=1}^{m}\mathcal{C}_j.
\end{equation}
We retain seed tasks for which
$\operatorname{Ans}(\Phi^{(0)})=\{x^\star\}$.
The formal representation records the target, supporting relations, intermediate entities, and the constraints required to recover the answer, providing an executable specification for subsequent expansion and validation.

\paragraph{Layer-wise Expansion.}
Starting from the seed representation, we iteratively increase the task's search depth.
At expansion step $\ell$, the synthesizer selects an expandable leaf constant $c$ from $\Phi^{(\ell)}$ and treats it as the answer to a new subproblem.
The subproblem is represented by a predicate $\Phi_c$ satisfying
$\operatorname{Ans}(\Phi_c)=\{c\}$.
We then replace the original constant with a new variable $z$ constrained by this subproblem:
\begin{equation}
    \Phi^{(\ell+1)}(x)
    =
    \exists z\,
    \left[
        \Phi^{(\ell)}(x;c\leftarrow z)
        \land
        \Phi_c(z)
    \right],
\end{equation}
where $\Phi^{(\ell)}(x;c\leftarrow z)$ denotes the expression obtained by replacing $c$ with $z$ in the current task representation.
Consequently, information that was directly available in the seed task must now be recovered through an additional search process.
The expansion is accepted only when the updated representation preserves the original target:
\begin{equation}
    \operatorname{Ans}\!\left(\Phi^{(\ell+1)}\right)
    =
    \{x^\star\}.
\end{equation}
Repeated expansion produces chained and branching structures whose intermediate results are necessary for resolving the final answer.

Each expansion combines relations from the knowledge graph with information retrieved from the Web.
The synthesizer searches for the selected leaf entity using multiple query formulations and collects relevant facts from heterogeneous pages.
We prioritize independently supported information and remove sources that merely duplicate the same underlying statement.
The retrieved facts are linked back to the corresponding entities before being incorporated into the formal representation, reducing errors caused by ambiguous names or mismatched entities.
When a Web document contains descriptive text instead of an explicit structured relation, candidate entities are first extracted and linked before being used as new expansion anchors.

\paragraph{Controlling Task Complexity.}
We separately control structural complexity and clue specificity.
Structural complexity is determined by the expansion depth, relation-chain length, number of branches, and number of constraints.
The resulting tasks cover several composition patterns, including chained retrieval, multi-constraint intersection, comparison, numerical reasoning, and temporal reasoning.

After the structure is fixed, we vary clue specificity through controlled transformations.
Exact dates may be converted into truthful time ranges, numerical values into intervals, entity names into type-level descriptions, and directly searchable attributes into indirect relational descriptions.
For example, an exact award name and year may be expressed through its field, issuing organization, and approximate period.
The original and transformed values are retained in the task metadata so that each rewritten clue can be checked against its underlying fact.

We vary the degree of transformation across constraints.
Some clues provide viable entry points for search, while others require broader exploration and cross-document integration.
This produces tasks with different search horizons without removing the information required to identify the answer.

The expanded formal representation is finally verbalized into a natural-language question.
The generator is instructed to preserve all factual constraints, avoid exposing the masked target, and express the clues as a coherent information-seeking request.
Each generated sample includes the question, reference answer, formal representation, supporting relations, source provenance, and transformation metadata.

\paragraph{Verification and Filtering.}
We apply several complementary filters before using a synthesized task for trajectory collection.
First, structural validation checks that the formal representation is internally consistent, contains no unresolved or cyclic dependencies, and preserves the reference answer throughout expansion.
Second, evidence validation examines the reliability and relevance of the supporting Web pages, verifies entity correspondence, and confirms that the reference answer satisfies every generated constraint.
Sources that are factually unsupported, low quality, redundant, or inconsistent with the target entity are removed together with the corresponding samples.

We approximate answer uniqueness using a targeted candidate set.
For a generated question $q$, we construct
\begin{equation}
    \mathcal{N}(q)
    =
    \mathcal{N}_{\mathrm{KG}}(q)
    \cup
    \mathcal{N}_{\mathrm{type}}(q)
    \cup
    \mathcal{N}_{\mathrm{agent}}(q),
\end{equation}
where $\mathcal{N}_{\mathrm{KG}}$ contains nearby entities sampled from the knowledge graph, $\mathcal{N}_{\mathrm{type}}$ contains entities of the same semantic type as the reference answer, and $\mathcal{N}_{\mathrm{agent}}$ contains plausible alternative answers produced during search-agent attempts.
Each candidate $x\in\mathcal{N}(q)$ is checked against the complete constraint set.
We discard a sample if any alternative candidate also satisfies its formal representation:
\begin{equation}
    \exists x\in
    \mathcal{N}(q)\setminus\{x^\star\}
    \quad
    \text{s.t.}
    \quad
    \Phi(x)=1.
\end{equation}

Finally, we evaluate contamination and empirical difficulty.
Tool-free models are first asked to answer each question directly, and questions that can be reliably solved without retrieval are removed.
For the remaining samples, a search agent performs $K$ independent attempts, yielding an empirical success rate
\begin{equation}
    \widehat{p}(q)
    =
    \frac{1}{K}
    \sum_{k=1}^{K}
    \mathbb{I}
    \left[
        \hat{y}_k=x^\star
    \right].
\end{equation}
We use this estimate to remove tasks that are consistently trivial or effectively unsolvable and to construct a difficulty-balanced training set.
The retained tasks are subsequently executed with our agent harness to collect complete trajectories for supervised fine-tuning and reinforcement learning.

\section{Qualitative Analysis of Autonomous Context Recovery}
\label{sec:seal-cases}

We further examine whether the model uses context management as an active recovery mechanism, rather than invoking it only when the context window is nearly exhausted. We identify successful trajectories in which the model explicitly recognizes that its current search is no longer productive, autonomously invokes \texttt{Seal Memory}, and subsequently changes its search strategy. Table~\ref{tab:seal-cases} summarizes two representative cases.

\begin{table*}[t]
\centering
\small
\setlength{\tabcolsep}{2pt}
\renewcommand{\arraystretch}{1.15}
\caption{Representative trajectories exhibiting autonomous context recovery. In both cases, the model invokes \texttt{Seal Memory} without an externally imposed compaction trigger, preserves the useful search state, and changes its strategy after entering a refreshed context.}
\label{tab:seal-cases}
\begin{tabular}{@{}p{0.12\textwidth}p{0.22\textwidth}p{0.24\textwidth}p{0.26\textwidth}p{0.11\textwidth}@{}}
\toprule
Case & Before Seal & Sealed State & Strategy after Reset & Outcome \\
\midrule
Hipolit Łossowski
&
The model repeatedly examines Polish officers, but each candidate violates at least one constraint, such as military branch, award rank, death place, or burial site.
&
It records the rejected candidates together with their exclusion reasons, while retaining the unresolved constraints and verified historical facts.
&
The model replaces person-by-person enumeration with a structured query combining war participation, military rank, award, death place, and burial location.
&
1881 \checkmark
\\
\addlinespace
Hiroaki Yamamoto
&
The model searches the current Rakuten Eagles roster but cannot find a player satisfying both the birth-year and weight constraints, eventually noting that it is ``running in circles.''
&
It preserves the likely team, the failure of the current-roster hypothesis, conflicting weight evidence, and the possibility that the target is a former player.
&
The model re-verifies the weight constraint, expands the search from the current roster to historical players, and corrects its structured-query formulation.
&
Hiroaki Yamamoto \checkmark
\\
\bottomrule
\end{tabular}
\end{table*}

In the first case, the model initially searches for a Polish military officer by checking individual candidates. This produces a sequence of partial matches: some candidates participated in the relevant war but received the wrong class of the Order of Polonia Restituta, while others fail the death-place or burial constraint. After recognizing that continued enumeration is inefficient, the model invokes \texttt{Seal Memory} and stores not only the verified facts but also the rejected candidates and their exclusion reasons. In the refreshed segment, these dead ends are converted into explicit query constraints. The model then performs a joint structured search over military service, award rank, death place, and burial location, identifies Hipolit Łossowski, and returns the correct birth year, 1881.

The second case illustrates recovery from an incorrectly bounded search space. While identifying a Japanese baseball player, the model initially assumes that the target must appear on the current Rakuten Eagles roster. Repeated searches fail to reconcile the required birth year and weight, leading the model to state, ``I'm running in circles. Let me try a completely different approach.'' It then invokes \texttt{Seal Memory}, recording that the current-roster assumption has failed and that historical players should be considered. After the reset, the model verifies the disputed weight condition, expands the search to former Rakuten players, and corrects an entity-type error in its structured query. This revised search identifies Hiroaki Yamamoto and produces the correct answer.

These cases show that the learned behavior goes beyond periodic context compression. The model can recognize an unproductive search state, decide when to create a semantic boundary, preserve evidence and failed hypotheses, and redirect the subsequent search from a refreshed context. The controlled comparison with automatic context compaction provides complementary quantitative evidence for the effectiveness of this model-initiated recovery behavior.

\section{Reward Shaping and Context-Budget Randomization}
\label{app:reward-shaping}
Answer correctness provides the primary trajectory-level signal. We retain both positive and negative values of the group-relative advantage $\widehat{A}_i$ defined above, and augment this outcome signal with resource penalties and action-local signals on the final segment. We denote the correctness component assigned to each trainable token in the final segment by $A_i^{\mathrm{acc}}=\widehat{A}_i$.

For resource usage, let $N_i$ be the number of assistant turns, $T_{\max}$ the turn limit, $L_i$ the length of the active segment, and $B_{\mathcal{G}}$ its assigned context budget. We use linearly increasing penalties near the corresponding limits:
\begin{align}
    p_i^{\mathrm{turn}}
    &=
    -\min\left(
        1,\,
        \frac{
            [N_i-(1-\rho_{\mathrm{turn}})T_{\max}]_{+}
        }{
            \rho_{\mathrm{turn}}T_{\max}
        }
    \right), \\
    p_i^{\mathrm{ctx}}
    &=
    -\min\left(
        1,\,
        \frac{
            [L_i-(1-\rho_{\mathrm{ctx}})B_{\mathcal{G}}]_{+}
        }{
            \rho_{\mathrm{ctx}}B_{\mathcal{G}}
        }
    \right),
\end{align}
where $[x]_{+}=\max(x,0)$. Thus, the turn penalty begins when a trajectory enters the final $\rho_{\mathrm{turn}}$ fraction of its turn budget, which is set to $5\%$ in our experiments. The context penalty follows the same pattern as the active segment approaches $B_{\mathcal{G}}$, and generation is terminated once the hard context limit is reached.

Format supervision is assigned to the assistant turn that produces the error. If turn $t$ contains $e_{i,t}$ format violations, its penalty is
\begin{equation}
    p_{i,t}^{\mathrm{fmt}}
    =
    -\min\left(
        \eta_{\mathrm{fmt}}e_{i,t},
        c_{\mathrm{fmt}}
    \right),
\end{equation}
where $\eta_{\mathrm{fmt}}$ controls the penalty per error and $c_{\mathrm{fmt}}$ limits its magnitude. We also identify three deterministic action violations: fabricated tool names or arguments, repeated identical search actions, and answer-ready \textsc{Seal Memory} calls. The last case occurs when the Seal arguments indicate that the model has already obtained the answer; the Seal operation is rejected and the model continues from the current segment. For an offending action of type $k$, we cap its token-level advantage by
\begin{equation}
    A_{i,t}
    \leftarrow
    \min\left(A_{i,t},-c_k\right),
\end{equation}
where $c_k$ is the corresponding violation-specific threshold.

We additionally reward parallel tool use when it reduces serial search rounds. Among correct trajectories for the same question, we compute the tool-use cost
\begin{equation}
    C_i
    =
    N_i^{\mathrm{round}}
    +
    \lambda_{\mathrm{call}}N_i^{\mathrm{call}},
\end{equation}
and rank the trajectories according to $C_i$. More efficient correct trajectories receive a larger efficiency score, which is converted into a width-aware bonus and assigned directly to the assistant turns that issue parallel tool calls. Combining these terms gives the final token-level advantage
\begin{equation}
    \widetilde{A}_{i,t}
    =
    A_i^{\mathrm{acc}}
    +
    \lambda_{\mathrm{ctx}}p_i^{\mathrm{ctx}}
    +
    \lambda_{\mathrm{turn}}p_i^{\mathrm{turn}}
    +
    \lambda_{\mathrm{fmt}}p_{i,t}^{\mathrm{fmt}}
    +
    \lambda_{\mathrm{par}}b_{i,t}^{\mathrm{parallel}},
\end{equation}
followed by the deterministic action caps above. When a rollout group has identical correctness rewards, ordinary auxiliary advantages are suppressed, while the resource penalties and deterministic action constraints remain active.

During training, the segment budget is sampled independently for each rollout group:
\begin{equation}
    B_{\mathcal{G}}
    \sim
    \operatorname{Uniform}(\mathcal{B}),
\end{equation}
where $\mathcal{B}$ contains multiple context limits. All trajectories for the same question share $B_{\mathcal{G}}$, preserving a matched resource setting within the group. After \textsc{Seal Memory}, the new segment starts from the task specification and sealed memory under the same budget. Varying $B_{\mathcal{G}}$ across groups requires the model to determine when to Seal from its current progress and remaining token budget. This prevents the policy from learning a fixed Seal trigger tied to a particular context length.

\section{BrowseComp-Lite: A Lightweight Evaluation Subset}
\label{app:browsecomp-lite}

BrowseComp contains 1,266 questions and requires long-horizon interactions with Web search tools~\citep{wei2025browsecomp}. Evaluating every model checkpoint on the full benchmark is therefore computationally expensive. To support more efficient model development and ablation studies, we construct \textsc{BrowseComp-Lite-185}, a fixed subset of 185 questions selected to closely approximate full-set performance.

We first apply a fixed random permutation to the original BrowseComp test set and define the first $K$ questions as a candidate subset $\mathcal{B}_K$. We then retrospectively evaluate each candidate subset using historical full-set results from multiple models and training checkpoints. For each evaluation setting, we compare the accuracy on $\mathcal{B}_K$ with its corresponding accuracy on the complete benchmark and identify the smallest $K$ that satisfies a prescribed deviation tolerance. This calibration is conducted using five historical Best-of-1 evaluation runs and four Best-of-5 runs.

\begin{table}[H]
    \centering
    \small
    \setlength{\tabcolsep}{6pt}
    \begin{tabular}{lcccccc}
        \toprule
        & \# Runs & \multicolumn{5}{c}{Deviation tolerance (percentage points)} \\
        \cmidrule(lr){3-7}
        Setting & & $1$ & $2$ & $3$ & $4$ & $5$ \\
        \midrule
        Best-of-1 & 5 & 345 & 230 & 185 & --  & --  \\
        Best-of-5 & 4 & 785 & 695 & 650 & 190 & 185 \\
        \bottomrule
    \end{tabular}
    \caption{Minimum subset size $K$ required to satisfy different subset-to-full-set deviation tolerances on the historical calibration runs.}
    \label{tab:browsecomp-lite-calibration}
\end{table}

As shown in Table~\ref{tab:browsecomp-lite-calibration}, selecting 185 questions limits the observed deviation to three percentage points under Best-of-1 evaluation and five percentage points under Best-of-5 evaluation. We therefore fix $K=185$ and use the resulting subset as \textsc{BrowseComp-Lite-185}.

\clearpage
\section{Qualitative Comparison of Active Sealing and Automatic Compaction}

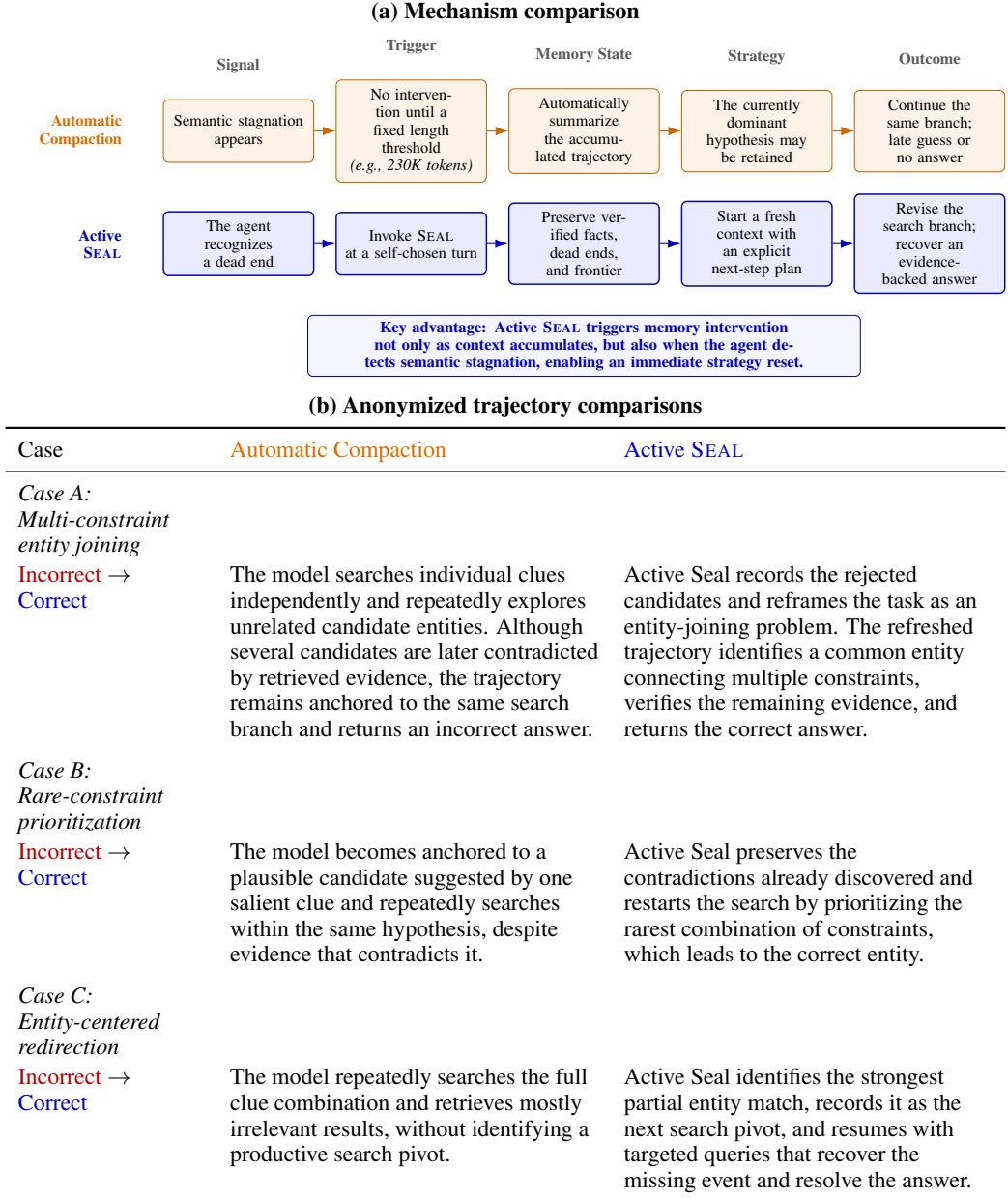
\begin{figure}[H]
    \centering
    {\small\bfseries (a) Mechanism comparison}\par
    \vspace{1mm}
    \resizebox{0.98\textwidth}{!}{%
    \begin{tikzpicture}[
        stage/.style={
            draw,
            rounded corners=3pt,
            text width=2.65cm,
            minimum height=1.25cm,
            align=center,
            font=\small,
            inner sep=5pt
        },
        passive/.style={
            stage,
            fill=orange!10,
            draw=orange!75!black
        },
        active/.style={
            stage,
            fill=blue!8,
            draw=blue!70!black,
            line width=0.8pt
        },
        flow/.style={
            -{Latex[length=2.6mm,width=1.8mm]},
            semithick
        },
        passiveflow/.style={
            flow,
            draw=orange!80!black
        },
        activeflow/.style={
            flow,
            draw=blue!75!black
        },
        header/.style={
            font=\small\bfseries,
            text=black!65,
            align=center
        },
        rowlabel/.style={
            font=\small\bfseries,
            text width=2.2cm,
            align=right
        }
    ]

    \node[passive] (p1) at (0,1.8)
        {Semantic stagnation\\appears};

    \node[passive] (p2) at (3.45,1.8)
        {No intervention until a\\fixed length threshold\\
        \textit{(e.g., 230K tokens)}};

    \node[passive] (p3) at (6.90,1.8)
        {Automatically summarize\\the accumulated trajectory};

    \node[passive] (p4) at (10.35,1.8)
        {The currently dominant\\hypothesis may be retained};

    \node[passive] (p5) at (13.80,1.8)
        {Continue the same branch;\\
        late guess or no answer};

    \draw[passiveflow] (p1) -- (p2);
    \draw[passiveflow] (p2) -- (p3);
    \draw[passiveflow] (p3) -- (p4);
    \draw[passiveflow] (p4) -- (p5);

    \node[active] (a1) at (0,-0.45)
        {The agent recognizes\\a dead end};

    \node[active] (a2) at (3.45,-0.45)
        {Invoke \textsc{Seal}\\
        at a self-chosen turn};

    \node[active] (a3) at (6.90,-0.45)
        {Preserve verified facts,\\
        dead ends, and frontier};

    \node[active] (a4) at (10.35,-0.45)
        {Start a fresh context with\\
        an explicit next-step plan};

    \node[active] (a5) at (13.80,-0.45)
        {Revise the search branch;\\
        recover an evidence-backed answer};

    \draw[activeflow] (a1) -- (a2);
    \draw[activeflow] (a2) -- (a3);
    \draw[activeflow] (a3) -- (a4);
    \draw[activeflow] (a4) -- (a5);

    \node[header, above=4mm of p1] {Signal};
    \node[header, above=4mm of p2] {Trigger};
    \node[header, above=4mm of p3] {Memory State};
    \node[header, above=4mm of p4] {Strategy};
    \node[header, above=4mm of p5] {Outcome};

    \node[rowlabel, left=7mm of p1, text=orange!85!black]
        {Automatic\\Compaction};

    \node[rowlabel, left=7mm of a1, text=blue!75!black]
        {Active\\\textsc{Seal}};

    \node[
        draw=blue!55!black,
        fill=blue!4,
        rounded corners=3pt,
        font=\small\bfseries,
        text=blue!75!black,
        align=center,
        text width=10.8cm,
        below=6mm of a3
    ] (takeaway) {
        Key advantage: Active \textsc{Seal} triggers memory intervention
        not only as context accumulates, but also when the agent detects
        semantic stagnation, enabling an immediate strategy reset.
    };

    \end{tikzpicture}%
    }

    \vspace{1mm}

    {\small\bfseries (b) Anonymized trajectory comparisons}\par
    \vspace{1mm}

    \begingroup
    \footnotesize
    \setlength{\tabcolsep}{5pt}
    \renewcommand{\arraystretch}{1.18}
    \begin{tabularx}{0.98\textwidth}{
        >{\raggedright\arraybackslash}p{2.55cm}
        >{\raggedright\arraybackslash}X
        >{\raggedright\arraybackslash}X}
        \toprule
        Case &
        \textcolor{orange!85!black}{Automatic Compaction} &
        \textcolor{blue!75!black}{Active \textsc{Seal}} \\
        \midrule

        \textit{Case A: Multi-constraint entity joining}\\
        \textcolor{red!70!black}{Incorrect}
        $\rightarrow$
        \textcolor{blue!75!black}{Correct}
        &
        The model searches individual clues independently and repeatedly
        explores unrelated candidate entities. Although several candidates
        are later contradicted by retrieved evidence, the trajectory remains
        anchored to the same search branch and returns an incorrect answer.
        &
        Active Seal records the rejected candidates and reframes the task as
        an entity-joining problem. The refreshed trajectory identifies a
        common entity connecting multiple constraints, verifies the remaining
        evidence, and returns the correct answer.
        \\

        \addlinespace[1.5mm]
        \textit{Case B: Rare-constraint prioritization}\\
        \textcolor{red!70!black}{Incorrect}
        $\rightarrow$
        \textcolor{blue!75!black}{Correct}
        &
        The model becomes anchored to a plausible candidate suggested by one
        salient clue and repeatedly searches within the same hypothesis,
        despite evidence that contradicts it.
        &
        Active Seal preserves the contradictions already discovered and
        restarts the search by prioritizing the rarest combination of
        constraints, which leads to the correct entity.
        \\

        \addlinespace[1.5mm]
        \textit{Case C: Entity-centered redirection}\\
        \textcolor{red!70!black}{Incorrect}
        $\rightarrow$
        \textcolor{blue!75!black}{Correct}
        &
        The model repeatedly searches the full clue combination and retrieves
        mostly irrelevant results, without identifying a productive search
        pivot.
        &
        Active Seal identifies the strongest partial entity match, records it
        as the next search pivot, and resumes with targeted queries that
        recover the missing event and resolve the answer.
        \\

        \bottomrule
    \end{tabularx}
    \endgroup

    \caption{
        Mechanism and qualitative comparison of automatic compaction
        and active \textsc{Seal}. (a) Automatic compaction is triggered by
        context length and can miss an earlier point of semantic stagnation;
        when triggered late, it may summarize an already entrenched search
        branch. Active \textsc{Seal} instead couples the agent's recognition
        of a dead end with an immediate, structured state transition.
        (b) Three anonymized cases retain only the behavioral differences
        between the two strategies: multi-constraint entity joining,
        rare-constraint prioritization, and entity-centered redirection.
        To avoid potential benchmark contamination, we anonymize the cases
        by removing task-specific entities, sources, dates, topics,
        quotations, clues, and final answers while retaining only
        behavior-level differences. 
    }
    \label{fig:active-seal-vs-auto-compaction}
\end{figure}

As illustrated in Figure~\ref{fig:active-seal-vs-auto-compaction},
the main distinction is not merely how the trajectory is compressed,
but when and why the state transition occurs. Automatic compaction
responds only to context length, while active \textsc{Seal} can respond
both to context consumption and to agent-recognized semantic stagnation.
\section{Tool Interfaces}
\label{app:tool-schemas}

\lstdefinestyle{toolschema}{
    basicstyle=\ttfamily\scriptsize,
    breaklines=true,
    breakatwhitespace=false,
    columns=fullflexible,
    keepspaces=true,
    showstringspaces=false,
    frame=single,
    framesep=3pt,
    rulecolor=\color{black!25},
    backgroundcolor=\color{black!2},
    xleftmargin=2pt,
    xrightmargin=2pt
}
The answer agent interacts with the environment through four function-calling
interfaces: web search, question-conditioned page summarization, active memory
sealing, and memory retrieval. The schemas below are the interfaces exposed to
the model. Backend choices and execution policies---including the search
engine, result-field filtering, domain blocking, retry limits, and Seal-count
limits---are fixed by the experimental configuration and are not model-visible
arguments. All top-level argument objects are strictly validated and reject
undeclared fields via \texttt{additionalProperties: false}. Runtime budget
annotations are appended by the framework after tool execution and are not part
of the semantic return schema.

\subsection{Web Interaction Tools}

\subsubsection{Web search: \texttt{search\_api}}

The search tool accepts a free-form query and an optional result count. The
search backend is selected by the evaluator rather than by the agent, which
keeps the model-facing interface identical across evaluation settings.

\begin{lstlisting}[style=toolschema]
{
  "type": "function",
  "function": {
    "name": "search_api",
    "description": "Call search API to execute web search queries",
    "parameters": {
      "type": "object",
      "properties": {
        "query": {
          "type": "string",
          "description": "Search query keywords or question"
        },
        "return_n": {
          "type": "integer",
          "description": "Number of results to return",
          "default": 10
        }
      },
      "required": ["query"],
      "additionalProperties": false
    }
  }
}
\end{lstlisting}

On success, the tool returns a ranked list. In our configuration, each result
retains the title, URL, textual description, and rank position.

\begin{lstlisting}[style=toolschema]
{
  "return": [
    {
      "title": "...",
      "url": "https://...",
      "description": "...",
      "position": 0
    }
  ],
  "status": "SUCCESS_SEARCH"
}
\end{lstlisting}

\subsubsection{Page summary: \texttt{link\_summary\_tool}}

This tool reads one or more URLs and extracts information conditioned on the
agent's current question. Allowing multiple URLs supports comparison without
changing the call structure.

\begin{lstlisting}[style=toolschema]
{
  "type": "function",
  "function": {
    "name": "link_summary_tool",
    "description": "Summarize the content of specified URL(s) based on a question",
    "parameters": {
      "type": "object",
      "properties": {
        "question": {
          "type": "string",
          "description": "The question to answer or summarization requirements"
        },
        "url": {
          "description": "URL(s) of the webpage(s) to summarize",
          "oneOf": [
            {"type": "string"},
            {
              "type": "array",
              "items": {"type": "string"}
            }
          ]
        }
      },
      "required": ["question", "url"],
      "additionalProperties": false
    }
  }
}
\end{lstlisting}

A successful response contains the synthesized answer and reader status. The
implementation may retry transient failures or fall back to a reader-plus-LLM
pipeline, but this recovery is transparent to the answer agent.

\begin{lstlisting}[style=toolschema]
{
  "return": {
    "linkreader": ["SUCCESS_LINKREADER"],
    "summary": "..."
  },
  "status": "SUCCESS_LINKSUMMARY"
}
\end{lstlisting}

\subsection{Memory Tools}

\subsubsection{Active checkpoint: \texttt{seal\_memory\_tool}}

The Seal tool creates a structured cognitive checkpoint. Instead of storing an
unstructured transcript alone, it asks the agent to distinguish verified,
conflicting, and partial facts; record visited paths and dead ends; retain
unexplored leads; and specify the first action after the context transition.
Only \texttt{next\_step\_plan} and \texttt{task\_progress} are mandatory at
the top level, allowing early checkpoints to remain partial.

\begin{lstlisting}[style=toolschema]
{
  "type": "function",
  "function": {
    "name": "seal_memory_tool",
    "description": "Mid-task cognitive checkpoint to compress current context into a reusable state",
    "parameters": {
      "type": "object",
      "properties": {
        "knowledge_graph": {
          "type": "array",
          "description": "Structured facts extracted so far",
          "items": {
            "type": "object",
            "properties": {
              "fact": {"type": "string"},
              "source_url": {"type": "string"},
              "status": {
                "type": "string",
                "enum": ["verified", "conflicting", "partial"]
              }
            },
            "required": ["fact", "status"]
          }
        },
        "navigation_state": {
          "type": "object",
          "description": "Topological map of the browsing session",
          "properties": {
            "visited_summary": {"type": "string"},
            "frontier_queue": {
              "type": "array",
              "items": {
                "type": "object",
                "properties": {
                  "target": {"type": "string"},
                  "reason": {"type": "string"}
                }
              }
            },
            "dead_ends": {
              "type": "array",
              "items": {"type": "string"}
            }
          },
          "required": [
            "visited_summary",
            "frontier_queue",
            "dead_ends"
          ]
        },
        "meta_learnings": {
          "type": "array",
          "items": {"type": "string"}
        },
        "next_step_plan": {
          "type": "string",
          "description": "First action in the new context window"
        },
        "task_progress": {
          "type": "string",
          "description": "Current progress toward the user goal"
        },
        "stage": {"type": "string"},
        "tags": {
          "type": "array",
          "items": {"type": "string"}
        },
        "comment": {"type": "string"},
        "structured_summary": {"type": "string"}
      },
      "required": ["next_step_plan", "task_progress"],
      "additionalProperties": false
    }
  }
}
\end{lstlisting}

The tool returns a persistent memory identifier together with the normalized
checkpoint. The framework also records the timestamp and may attach the source
conversation internally.

\begin{lstlisting}[style=toolschema]
{
  "memory_id": "<uuid>",
  "timestamp": "<UTC timestamp>",
  "stage": "...",
  "tags": ["..."],
  "comment": "...",
  "knowledge_graph": [
    {
      "fact": "...",
      "source_url": "https://...",
      "status": "verified"
    }
  ],
  "navigation_state": {
    "visited_summary": "...",
    "frontier_queue": [
      {"target": "...", "reason": "..."}
    ],
    "dead_ends": ["..."]
  },
  "meta_learnings": ["..."],
  "next_step_plan": "...",
  "task_progress": "..."
}
\end{lstlisting}

\subsubsection{Checkpoint retrieval: \texttt{read\_memory\_tool}}

The retrieval tool expands the checkpoint produced by the immediately
preceding segment. The framework exposes only the identifier of this most
recent checkpoint to the model after a context transition. By default, the
tool returns the compact structured state only; the agent can explicitly
request bounded tails of the associated conversation or the optional
structured summary when additional detail is necessary.

\begin{lstlisting}[style=toolschema]
{
  "type": "function",
  "function": {
    "name": "read_memory_tool",
    "description": "Read the most recently sealed memory by its memory_id",
    "parameters": {
      "type": "object",
      "properties": {
        "memory_id": {
          "type": "string",
          "description": "UUID of the most recent checkpoint returned by seal_memory_tool"
        },
        "include_conversation": {
          "type": "boolean",
          "default": false
        },
        "max_messages": {
          "type": "integer",
          "default": 10
        },
        "max_chars": {
          "type": "integer",
          "default": 8000
        },
        "include_structured": {
          "type": "boolean",
          "default": false
        },
        "structured_max_chars": {
          "type": "integer",
          "default": 6000
        }
      },
      "required": ["memory_id"],
      "additionalProperties": false
    }
  }
}
\end{lstlisting}

The default response reproduces the compact checkpoint. When requested,
\texttt{conversation\_history} and \texttt{structured\_summary} are added
subject to the corresponding message and character limits.

\begin{lstlisting}[style=toolschema]
{
  "memory_id": "<uuid>",
  "timestamp": "<UTC timestamp>",
  "stage": "...",
  "tags": ["..."],
  "comment": "...",
  "knowledge_graph": [...],
  "navigation_state": {...},
  "meta_learnings": [...],
  "next_step_plan": "...",
  "task_progress": "...",
  "conversation_history": [...],
  "structured_summary": "..."
}
\end{lstlisting}

\subsection{Judge Models and Evaluation Protocols}
\label{app:judge_settings}

Table~\ref{tab:judge_settings} summarizes the judge models and evaluation protocols used for results produced with our harness. We use DeepSeek-V4-Flash to evaluate BC, BC-ZH, GAIA, and xbench with benchmark-specific answer-matching prompts. For DeepSearchQA and WideSearch, we follow their official evaluation implementations, including the prescribed judge models and scoring procedures.

\begin{table}[H]
\centering
\small
\caption{Judge models and evaluation protocols used in our experiments.}
\label{tab:judge_settings}
\setlength{\tabcolsep}{3pt}
\begin{tabularx}{\textwidth}{@{}>{\raggedright\arraybackslash}p{2.05cm}>{\raggedright\arraybackslash}p{2.85cm}>{\raggedright\arraybackslash}X>{\raggedright\arraybackslash}p{1.35cm}@{}}
\toprule
Benchmark & Judge Model & Evaluation Protocol & Metric \\
\midrule
BC
    & DeepSeek-V4-Flash
    & Semantic equivalence between the predicted and reference answers
    & Accuracy \\
BC-ZH
    & DeepSeek-V4-Flash
    & The same answer-equivalence protocol as BC, with Chinese questions and answers
    & Accuracy \\
GAIA
    & DeepSeek-V4-Flash
    & Benchmark-specific answer matching covering numerical, textual, list, and date equivalence
    & Accuracy \\
xbench
    & DeepSeek-V4-Flash
    & Chinese answer-extraction and equivalence prompt following the benchmark grading format
    & Accuracy \\
DeepSearchQA
    & Gemini-2.5-Flash
    & Official component-level evaluation for single- and set-valued answers~\citep{gupta2026deepsearchqabridgingcomprehensivenessgap}
    & Macro-F1 \\
WideSearch
    & GPT-4.1-2025-04-14
    & Official table parsing, normalization, semantic alignment, and column-level grading~\citep{wong2025widesearchbenchmarkingagenticbroad}
    & Item-F1 \\
\bottomrule
\end{tabularx}
\end{table}

\paragraph{Additional Benchmark Details.}
For FinSearchComp, we evaluate on a time-stable subset of the official 635-question release. Since the 244 time-sensitive questions are associated with answers fixed at the dataset's 2025 snapshot, evaluating them with a live search engine may incorrectly penalize up-to-date answers. We therefore exclude these questions and retain all 391 time-stable questions; this subset is not obtained through random sampling. For ShoppingComp, we report VPR, defined as the proportion of evaluation instances for which the agent produces a valid product-oriented response, and SoP (\emph{Satisfaction of Products}), which measures the average proportion of user-requirement rubrics satisfied by the recommended products.

\paragraph{BC and BC-ZH.}
We use the same judge template for BC and BC-ZH, retaining the original language of the question, reference answer, and model response. The judge returns a binary decision in JSON format. The complete prompt is shown below.

{\footnotesize
\begin{verbatim}
Please determine whether the provided answer is correct.

Question: <question>

Ground truth: <ground_truth>

User answer: <model_answer>

Evaluate whether the user answer is equivalent to the
ground truth. Consider:
1. Numeric equality, ignoring formatting differences.
2. Semantic equivalence, ignoring casing, punctuation,
   and whitespace.
3. For lists, whether the elements match regardless of order.
4. Equivalence between different date formats.

Ignore format inconsistency.

Return the result in JSON:
{
    "reasoning": "Brief explanation",
    "decision": true/false
}

Return JSON only, with no extra commentary.
\end{verbatim}
}

\paragraph{DeepSearchQA and WideSearch.}
For DeepSearchQA, we use the official Gemini-2.5-Flash judge prompt. Each expected answer component is evaluated independently, while unsupported additional answers are counted as false positives when computing Macro-F1. For WideSearch, we use the official GPT-4.1 evaluator. Model outputs are first parsed as structured tables and normalized using the benchmark-provided string, numerical, date, and URL processing rules. GPT-4.1 is invoked for semantic primary-key alignment and columns requiring LLM-based grading, after which Item-F1 is computed using the official aggregation procedure.

\end{document}